\documentclass[letterpaper, 10pt, conference]{ieeeconf}
\IEEEoverridecommandlockouts                              
\usepackage{amsmath}
\usepackage[noadjust]{cite}
\usepackage{graphicx}
\usepackage[bookmarks=true,colorlinks=true]{hyperref}
\usepackage{stfloats}
\usepackage[caption=false,font=footnotesize]{subfig}
\usepackage{xcolor}

\title{\LARGE \bf Probabilistic Collision Risk Estimation for Pedestrian Navigation}

\author{Amine Tourki$^{1,4}$ and Paul Prevel$^{1}$ and Nils Einecke$^{2}$ and Tim Puphal$^{2,3}$ and Alexandre Alahi$^{4}$
\thanks{$^{1}$Biped Robotics SA, 1015 Lausanne, Switzerland
        {\tt firstname@biped.ai}}%
\thanks{$^{2}$Honda Research Institute Europe GmbH, 63073 Offenbach, Germany,
        {\tt firstname.lastname@honda-ri.de}}
\thanks{$^{3}$Honda Research Institute Japan Co., Ltd., Wako, 351-0114 Saitama, Japan.
        {\tt firstname.lastname@honda-ri.de}}
\thanks{$^{4}$\'Ecole Polytechnique F\'ed\'erale de Lausanne, 1015 Lausanne, Switzerland
        {\tt firstname.lastname@epfl.ch}}%
}

\begin{document}
\maketitle
\thispagestyle{empty}
\pagestyle{empty}

\begin{abstract}
Intelligent devices for supporting persons with vision impairment are becoming more widespread, but they are lacking behind the advancements in intelligent driver assistant system. To make a first step forward, this work discusses the integration of the risk model technology, previously used in autonomous driving and advanced driver assistance systems, into an assistance device for persons with vision impairment. The risk model computes a probabilistic collision risk given object trajectories which has previously been shown to give better indications of an object's collision potential compared to distance or time-to-contact measures in vehicle scenarios. In this work, we show that the risk model is also superior in warning persons with vision impairment about dangerous objects. Our experiments demonstrate that the warning accuracy of the risk model is 67\% while both distance and time-to-contact measures reach only 51\% accuracy for real-world data.
\end{abstract}

\section{Introduction}
A recent study \cite{WHO_2019} from the World Health Organization estimates that 43 million people worldwide are living with blindness, while 295 million suffer from moderate to severe visual impairment. Individuals with vision impairment face several dangers that include low-hanging obstacles, stairs without tactile markers or navigating through crowded areas. Traditional tools such as white canes \cite{Kahaki_2023} or guide dogs \cite{Silverman_2022} remain the most trustworthy for navigation purposes. However, accidents such as head-level injuries are still common \cite{Manduchi_2011} and happen as often as once per month. These seemingly minor accidents can lead to serious injuries and erode confidence in navigating public spaces.

\begin{figure}[t]
    \centering
    \includegraphics[width=0.50\linewidth]{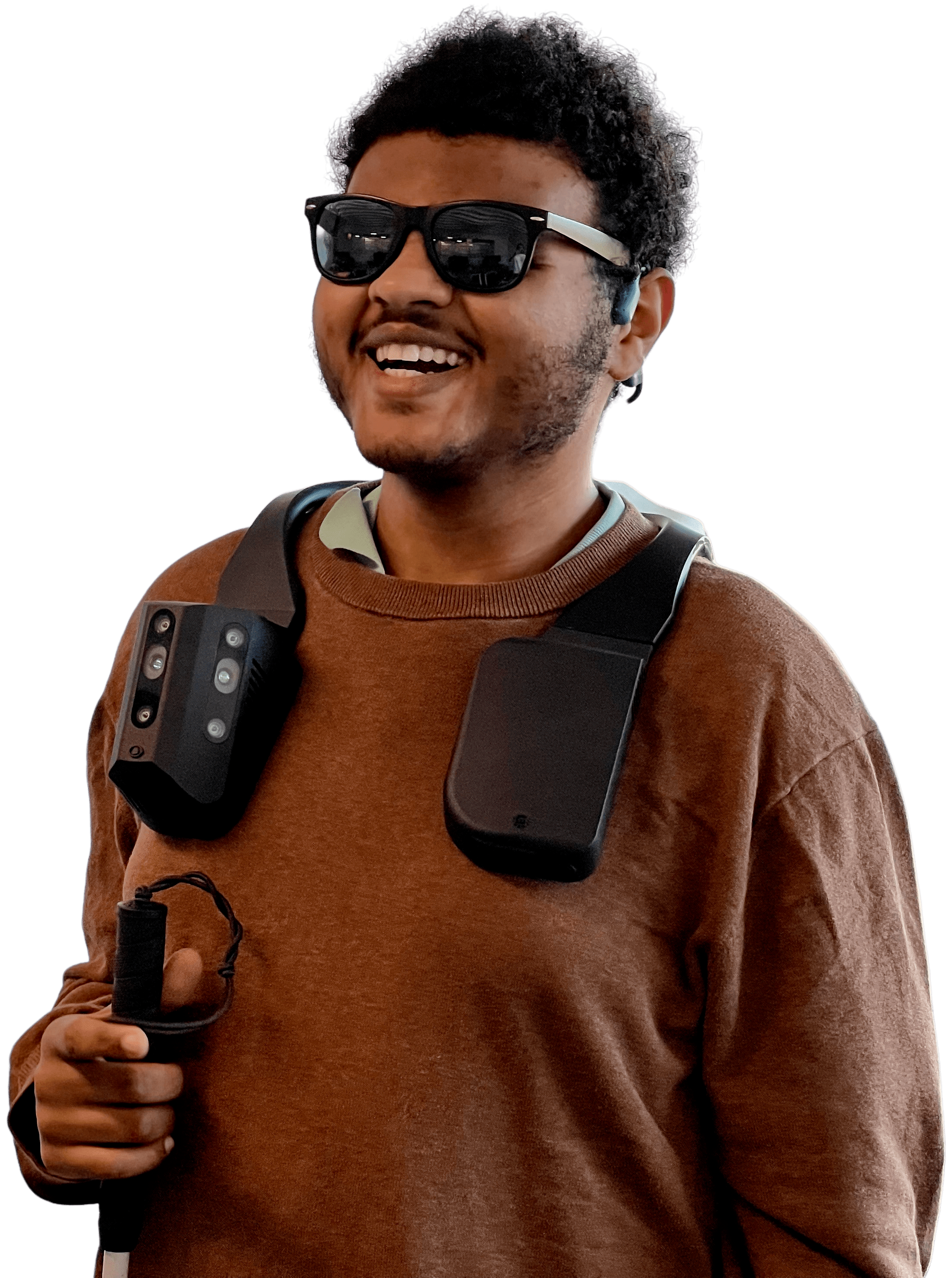}
    \caption{NOA assist device from Biped Robotics supporting people with vision impairment in navigation tasks. NOA stands for Navigation, Obstacle and AI.}
    \label{fig:Ali}
\end{figure}

New intelligent, camera-based devices \cite{Real_2019, Sessner_2022, Pittet_2025} are designed to detect and alert users of obstacles in real time. By acting as an extra set of eyes, these devices are transforming personal safety, offering enhanced independence and reducing the risk of falls and head-level injuries. One example for such a device is NOA (Navigation, Obstacle and AI.), developed by Biped Robotics~\cite{Pittet_2025}. As shown in Fig.~\ref{fig:Ali}, NOA is a wearable shoulder vest designed to assist the navigation of people with vision impairment. The device perceives the environment through camera sensors and runs a vision processing pipeline for object detection.

Assist devices, like NOA, support people with vision impairment in environment sensing, collision avoidance and navigation. These capabilities are similar to driver assistant systems or mobile robots. Algorithms for both are based on many years of research and development, hence it is a logical step to apply these advancements to the assist devices for people with vision impairment. One well-established technique from the intelligent car domain is the risk model presented in \cite{Eggert_2014, Puphal_2019}. This probabilistic collision risk estimation model has been applied successfully to autonomous driving \cite{Puphal_2018}, driver assistance systems \cite{Puphal_2022} and mobile robots~\cite{Honda_2024}.

In this work, we discuss the application of the risk model to pedestrian scenarios. Using synthetic data generated with Webots \cite{Michel_2004} and real-world data recorded with the NOA device, we demonstrate the superior accuracy of the risk model for collision warning compared to distance-based and time-to-contact-based approaches. While both distance and time-to-contact reach only 51\% accuracy for real-world data, the warning accuracy of the risk model is 67\%.

\section{Related works}
According to Real and Araujo \cite{Real_2019}, the first electronic assist devices for persons with vision impairment were based on sonar and radar sensors. These sensors were initially favored due to their simplicity in measuring distances, unlike camera-based and laser-based systems, which rely on costly and less precise triangulation. Later devices included GNSS receivers and concentrated more on the navigation aspects of assistance. At the beginning of the 2000s, cameras began to gain traction as they became smaller and more affordable. Cameras have the major advantage to be a potential replacement for impaired or lost vision. With the advent of smartphones they became popular devices for assisting persons with vision impairment. Unfortunately, most smartphone-based solutions in research and development are not aligned with the needs of users with vision impairment \cite{Budrionis_2022}. Thus, several new devices have been developed: smart glasses \cite{Bai_2017, Mukhiddinov_2021, Amore_2023}, augmented white canes \cite{Khan_2018, Messaoudi_2020, Mai_2023} and smart harnesses \cite{Bahadir_2012, Sessner_2022, Pittet_2025}. Many of these devices use cameras as their primary sensor and are designed to better accommodate the daily needs of users with vision impairment. However, their sensing software does not reach the level of driver assistance systems or mobile robots.

A field closely related to people navigation is the navigation of mobile robots through crowds \cite{Kruse_2013, Gao_2022}. For doing so, the robots need to predict the trajectory of several humans. As those trajectories are mutually depending on each other \cite{Helbing_1995l, Alahi_2016}, predicting them becomes difficult. Based on the predicted trajectories, the robot is planning its way through the crowd \cite{Aoude_2013}. Newer approaches are based on reinforcement learning and directly map either raw sensor signals \cite{Tai_2017, Long_2018} or perceived agent states \cite{Everett_2018, Chen_2019} to robot actions. While robot navigation has made tremendous advancements, the algorithms are not directly applicable to vision assistance systems. One reason is the computational cost. Vision assist devices need to run for hours and host also further functions in parallel, prohibiting costly model computation requiring a multi-core desktop CPU or a GPU like many robot navigation algorithms do \cite{Tai_2017, Long_2018, Everett_2018, Chen_2019}. Another more drastic reason is the different coupling. Vision assistance systems need to warn a person while traversing but navigation algorithms for robots are directly coupled to the robots movement control. While the robot can process all of the environment data in parallel on its internal CPU, the warning for persons with vision impairment needs to be reduced to a minimum of information with the highest relevance in order to not overwhelm or confuse the person. 

The latter problem is also well-known in the area of driver assistance systems \cite{Bliss_2003, Biondi_2018, Hasenjager_2019}. If not designed thoughtfully, the warning of an Advanced Driver Assistance System (ADAS) can be useless or, even worse, distract or confuse the driver. Also, different drivers may prefer different warnings, thus, the warning needs to be personalizable.

The NOA device (see Fig.~\ref{fig:Ali}) used in this work employs bone conducting ear phones to warn the user about the most relevant obstacle, similar to the park assist sounds in vehicles. The warning can be configured based on user preferences. For example, some users prefer constant warnings when following a wall, while others prefer an initial warning but no continuous warning during wall following. The object detection pipeline is mainly based on processing  point cloud data generated from RGB-D sensors. Object trajectories are derived in a linear manner by comparing object positions from two consecutive time steps. In this work, we analyze to what extend state-of-the-art collision risk estimation methods used in ADAS and autonomous driving, can improve the warning accuracy of the NOA blind assist device. To this end, we integrate the risk model \cite{Eggert_2014, Puphal_2019}, a probabilistic risk estimation framework, into the processing pipeline of NOA.

\section{Setup}

\subsection{Sensors}\label{sub:sensors}
\begin{figure}[t]
    \centering
    \subfloat[NOA cameras]{
        \includegraphics[width=0.8\linewidth]{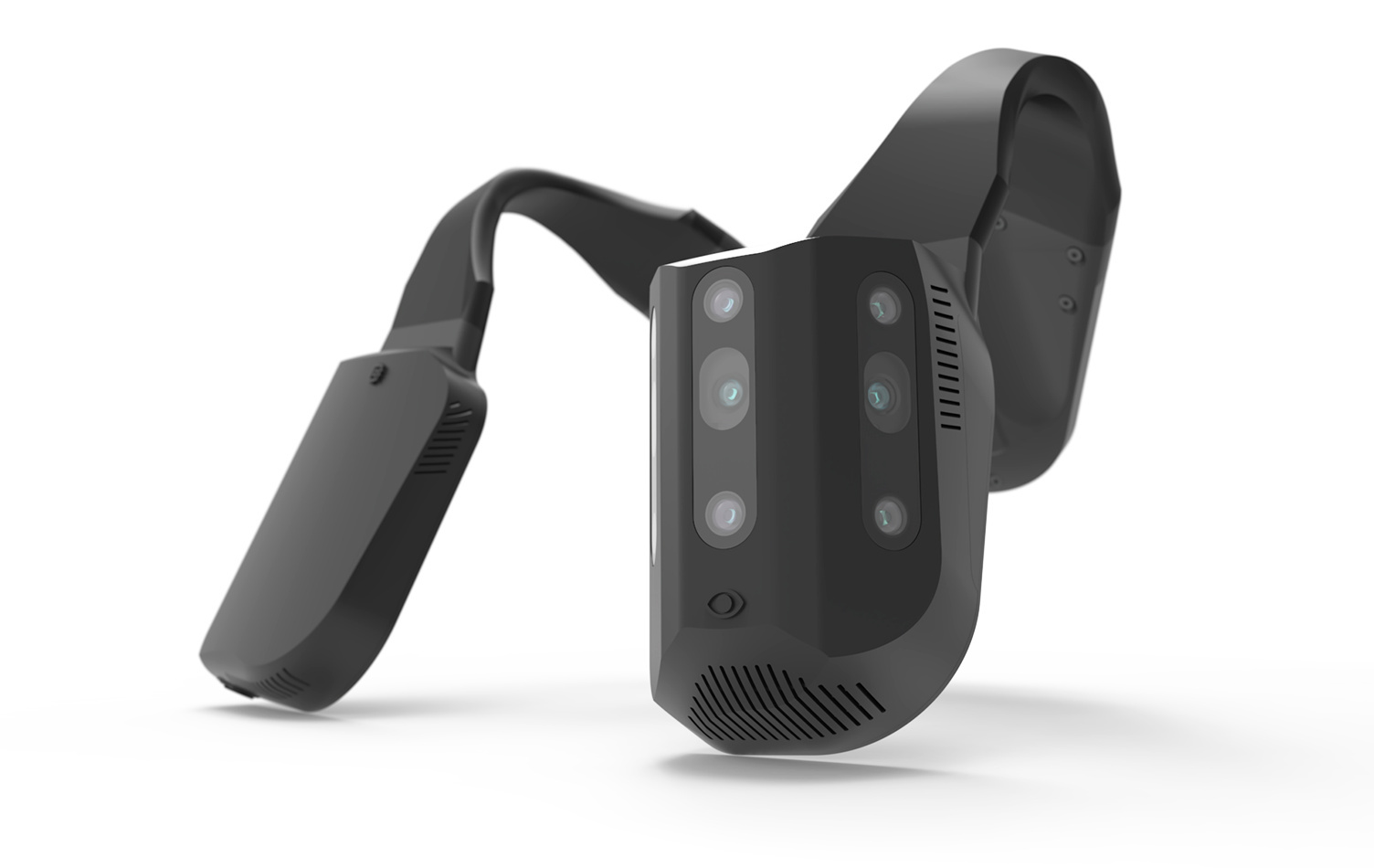}
        \label{fig:camera}
    }\\
    \subfloat[Stitched camera image]{
        \includegraphics[width=0.45\linewidth]{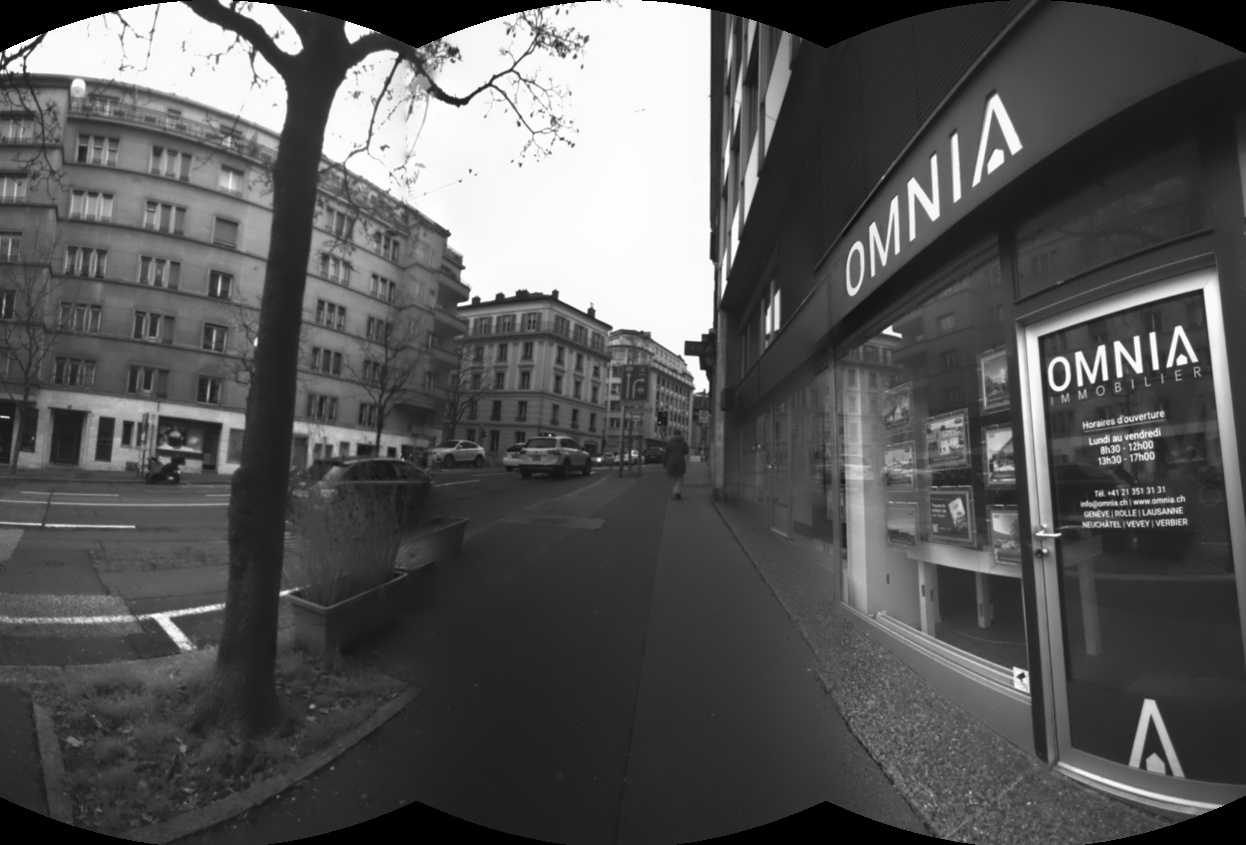}
        \label{fig:stitched_image}
    }
    \hfill
    \subfloat[3D obstacle segmentation]{
        \includegraphics[width=0.45\linewidth]{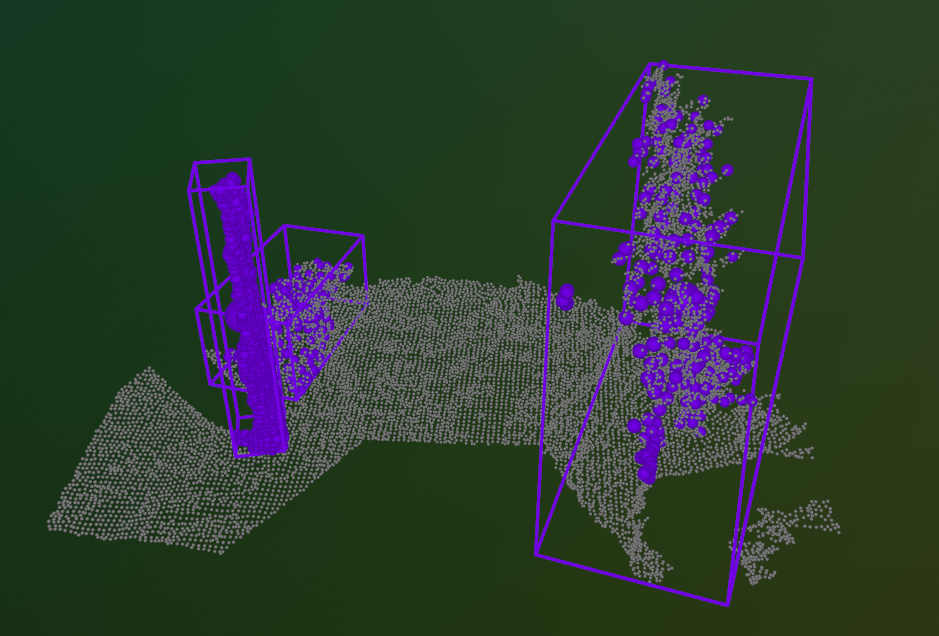}
        \label{fig:3d_segmentation}
    }
    \caption {\protect\subref{fig:camera} NOA camera setup. Three RealSense D430 are arranged in vertical orientation in a half-circle for a 170\textdegree\ horizontal and 90\textdegree\ vertical FoV. \protect\subref{fig:stitched_image} Example of stitching all three camera images into one coherent view. \protect\subref{fig:3d_segmentation} Point cloud and segmentation of scene shown in \protect\subref{fig:stitched_image}. Purple dots indicate object points and purple bounding boxes segmented objects.}
\end{figure}

The NOA assist device~\cite{Pittet_2025} uses RGB-D cameras as its main sensors, as shown in Fig.~\ref{fig:camera}. Three RealSense D430 

\noindent are arranged in vertical orientation in a half-circle, providing a 170\textdegree\ horizontal and a 90\textdegree\ vertical field of view (FoV). Fig.~\ref{fig:stitched_image} depicts a stitched scene image from the three cameras. The center camera is a RealSense D430i with an additional IMU. The camera depth maps are converted to a 3D scene point cloud for detecting objects. See Fig.~\ref{fig:3d_segmentation} for an example 3D point cloud with detected objects.

\subsection{Processing Pipeline}
\label{sub:copilot_pipeline}
\begin{figure*}[t]
    \centering
    \includegraphics[width=0.99\linewidth]{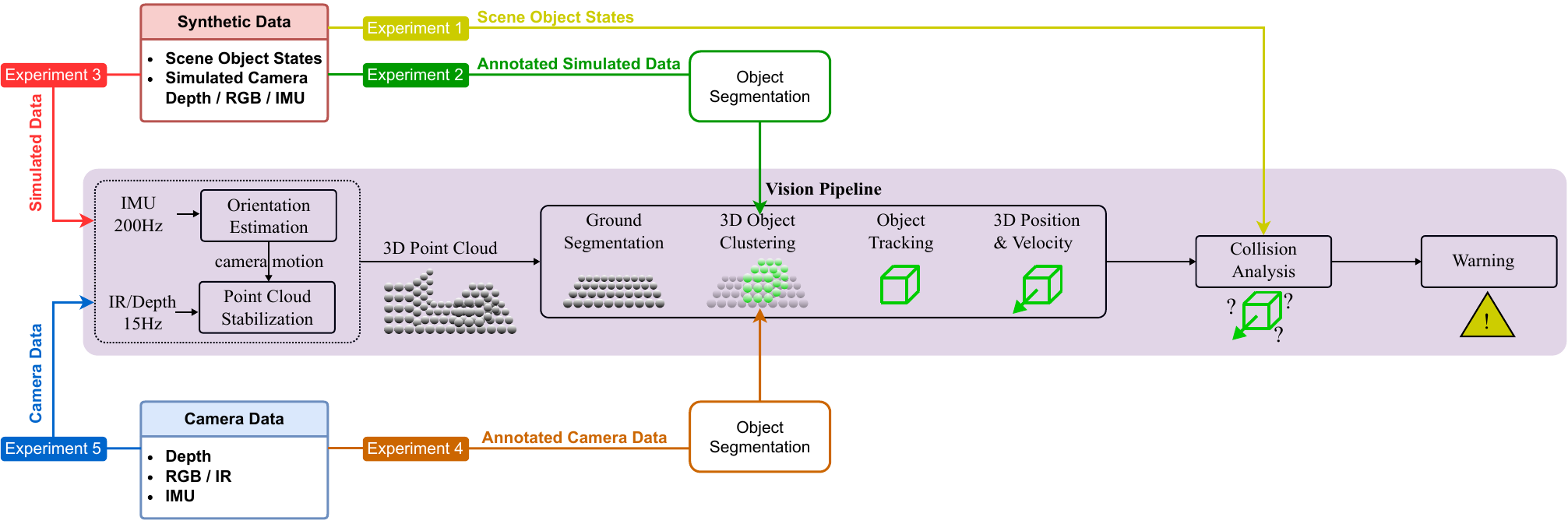}
    \caption{NOA processing pipeline (purple area) and experimental setups. The processing of the NOA device consists of aggregating a 3D scene point cloud, extracting the ground, cluster objects and tracking the objects. Each object trajectory is analyzed for crossing the user. In case of a detected future collision, the user is warned by auditory feedback. Five experimental setups are investigated in this work, for a detailed explanation please refer to section~\ref{sub:data}.}
    \label{fig:pipeline_with_experiments}
\end{figure*}

The main processing pipeline in NOA is depicted in the purple box in Fig.~\ref{fig:pipeline_with_experiments}. It starts with capturing data from the IMU and the cameras with 200 Hz and 15 Hz, respectively. The 3D point clouds generated by the three RealSense cameras are fused into a coherent scene point cloud. The IMU data is used to track the scene point cloud over time. Using the scene point cloud, the ground plane is estimated. This enables a separation of the point cloud into ground plane, holes and objects by means of 3D clustering. Objects are tracked over time by a Kalman Filter. Eventually, a 3D position and velocity is estimated for each object. This estimation is the basis for evaluating the collision of the user with the objects in the scene. An auditory warning with directional information is issued by the NOA device for imminent dangers. 

\subsection{Dataset}
\label{sub:data}

\begin{figure}
    \centering
    \subfloat[Side collision]{\includegraphics[width=0.45\linewidth]{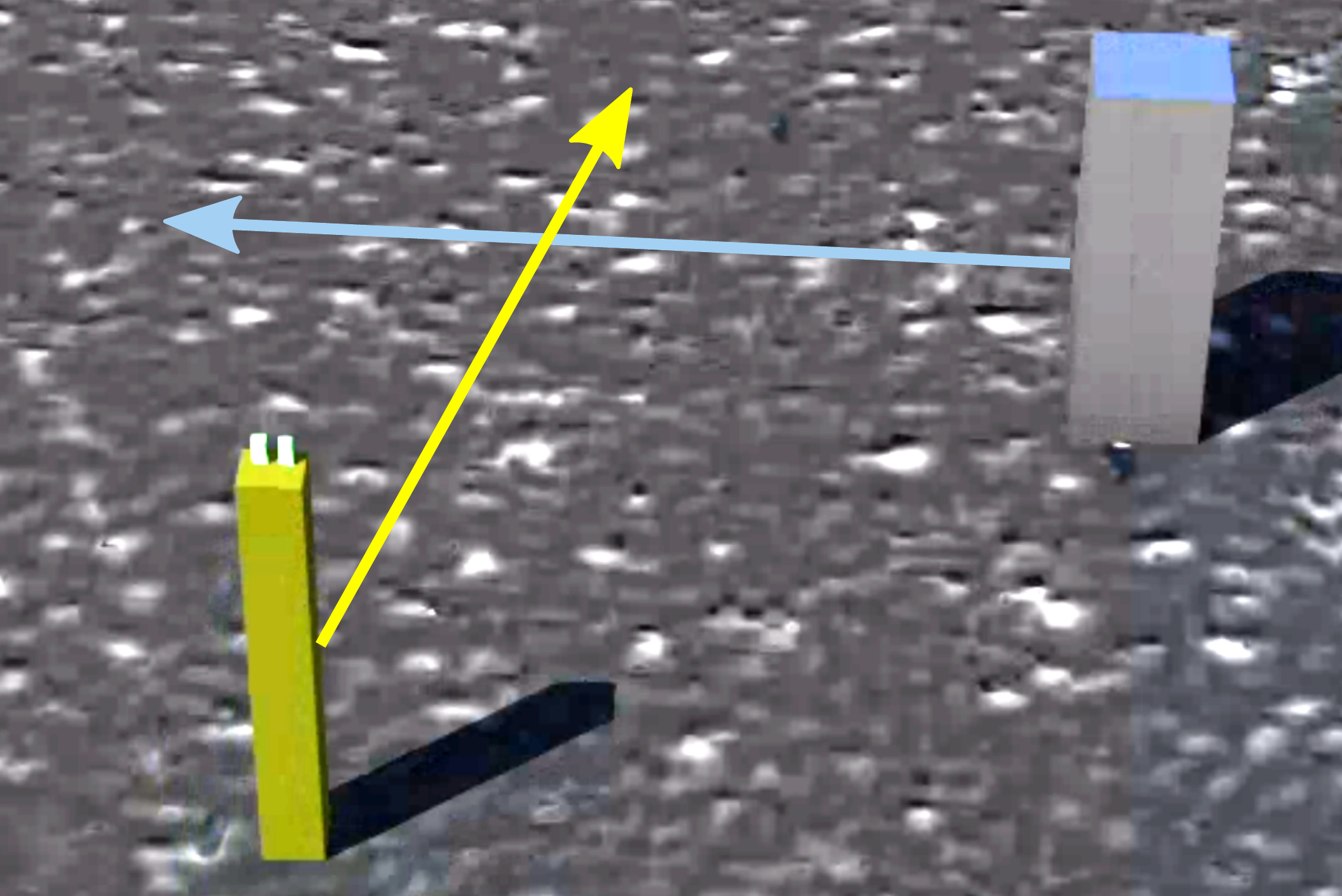}}
    \hfill
    \subfloat[Walking towards crowd]{\includegraphics[width=0.45\linewidth]{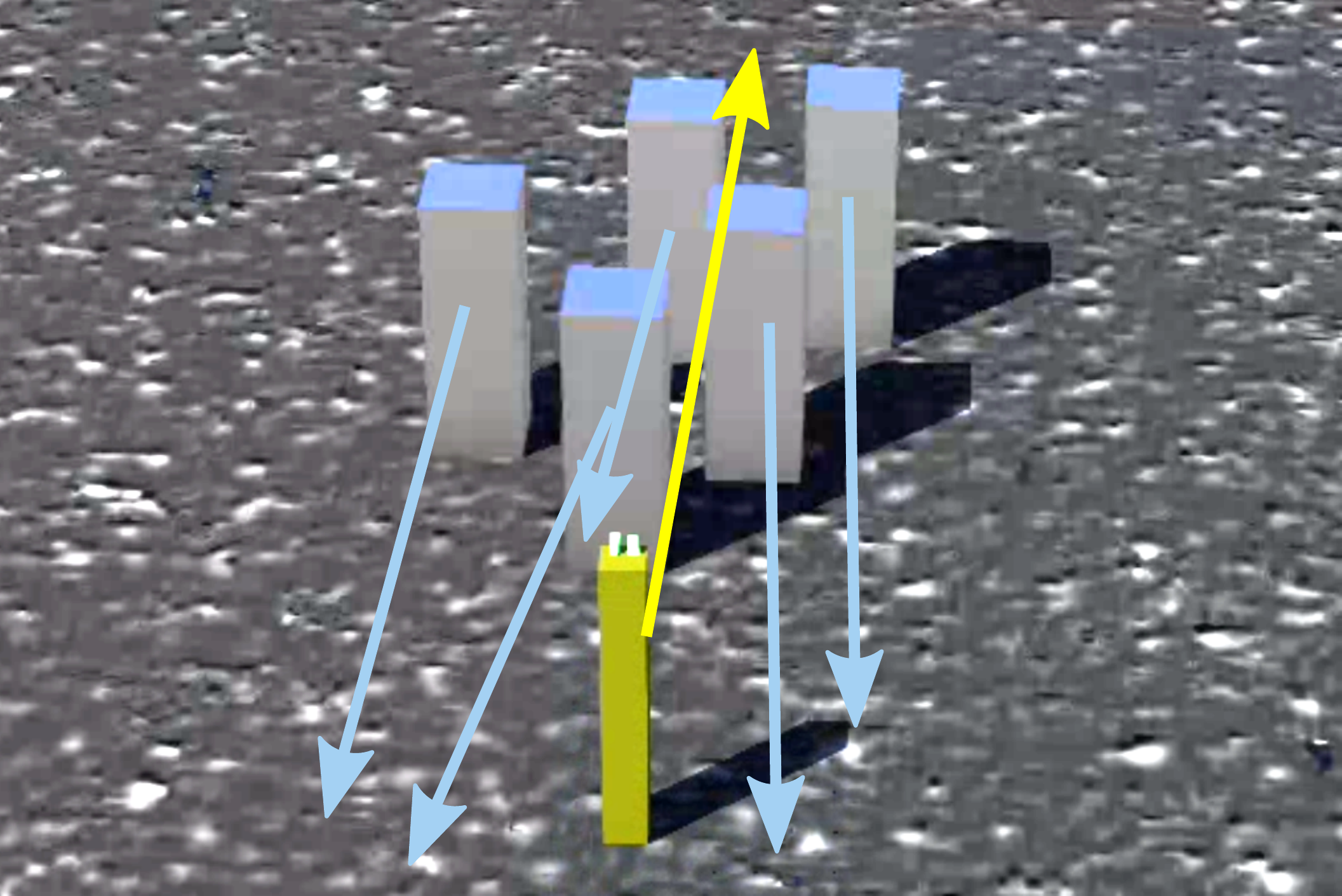}}
    \caption{Two example scenes used for generating simulated sensor data with Webots 3D robotics simulator.}
    \label{fig:simulated_data}
\end{figure}

In this work, we use two different datasets. The first dataset was recorded using the NOA device in various indoor and outdoor environments (sidewalks, university campus and train stations). It includes 30k frames captured along with IMU data. The second dataset was generated using Webots~\cite{Michel_2004}, a 3D robotics simulator. It includes 10k frames and consists of very specific scenarios such as side collisions or walking towards a crowd (see Fig.~\ref{fig:simulated_data}).

The real-world data set was recorded without orchestration. Testers, wearing the NOA device during recording, were asked to move naturally and recordings were done during different times of day and weather conditions. To reduce the data bias towards non-collision events, lengthy parts of the recordings without object interaction, like walking down an empty path, were removed for this data set. On average there are 1.7 objects in the synthetic scenes and 7.4 object in the real-world scenes, i.e., the real-world scenes are a lot busier. Furthermore, around 26\% of the objects in the synthetic scene are critical objects to warn about, while only 2\% of the objects in the real-world scene are critical.

The synthetic dataset was designed for testing with clean data without sensor noise. It includes rendered camera images as well as rendered depth maps and simulated IMU data. Furthermore, ground truth object positions and velocities, and ground truth image segmentation of the objects are available for the synthetic data. The movement of the user as well as the object trajectories were defined by hand. The trajectories encompass linear as well as non-linear movement. As the synthetic data set's main purpose is reference performance, it is limited to a smaller size. Like the real-world data set it is free of lengthy streams without object interaction, to concentrate on the performance of detecting and warning of user-object collisions. The warning performance in scenes with no or distant objects is mainly defined by correct detections of the NOA pipeline, which we do not want to asses here.

Based on these two datasets, we constructed five different experimental setups (see Fig.~\ref{fig:pipeline_with_experiments}) of different difficulty:

\textbf{\textcolor[HTML]{CCCC00}{Experiment 1 - Scene Object States:}} Ground truth object states (position, velocity) are used.

\textbf{\textcolor[HTML]{009900}{Experiment 2 - Annotated Simulated Data:}} Synthetic data together with object masks generated by the simulator software. 3D clustering is performed by projecting the 3D point cloud onto the 2D object masks, instead of using the onboard clustering of the NOA device.

\textbf{\textcolor[HTML]{FF3333}{Experiment 3 - Simulated Data:}} Synthetic data with simulated camera sensors is processed through the device full vision pipeline.

\textbf{\textcolor[HTML]{CC6600}{Experiment 4 - Annotated Camera Data:}} Real-world data together with hand-annotated object masks. 3D clustering is performed by projecting the 3D point cloud onto 2D object masks, instead of using the onboard clustering of the NOA device.

\textbf{\textcolor[HTML]{0066CC}{Experiment 5 - Camera Data:}} Real-world data is processed through the device full vision pipeline.

The different setups include different sources of noise and errors. While the sensor data in setups 1, 2 and 3 is noise-free, setups 4 and 5 include real-world sensor noise. Concretely, setup 1 is free of noise and free of obstacle state estimation errors, providing the opportunity to add artificial noise to analyze the warning algorithm's robustness against different noise sources. Please note, however, that the object state input to the algorithms consists only of position and velocity vector at each time step. The algorithms are not given the full object trajectories but always use a linear model to predict object movements. This can lead to prediction errors in case of non-linear object movement. In setups 2 and 4, the 3D clustering is based on labeled information, reducing errors in the vision pipeline to only tracking errors and trajectory prediction errors. Finally, setups 3 and 5 use the full vision processing pipeline of the NOA device and enable to analyze the real-world applicability. 

In all setups, the collision evaluation is provided with the scene object states. For each object, the state is represented with $\vec{s} = \begin{bmatrix} i & p_x & p_y & v_x & v_y \end{bmatrix}$, where $i$ is the tracking ID, $p_x$ and $p_y$ are the 2D coordinates, and $v_x$ and $v_y$ represent the velocity of the object. The coordinate system is a 2D space aligned with the ground plane, with its origin at the ego-position, meaning object positions and velocity are relative to the user. This also means that, from the algorithms point of view, the object movements are apparent movements that are superimposed with the movement of the user. The user movement is, so far, not estimated in the NOA pipeline.

\subsection{Ideal Warning}\label{sub:ideal_warning}
To analyze the performance of different collision evaluation methods, we first needed to understand what an ideal warning would look like from the perspective of a person with vision impairment. To this end, we conducted interviews with users with vision impairment and mobility trainers. We found considerable variation in the preferences of individuals. However, for a scientific and clear analysis of the warnings, we decided to find a compromise based on the opinions of those interviewed. This ideal warning consists of three key elements: 
\begin{enumerate}
    \item A warning should be given to the user at least 3 seconds before a potential collision with an object.
    \item An object is considered to lead to a potential collision if it enters a 1-meter radius "safety bubble" around the user, either immediately or anytime in the future.
    \item The warning should remain active until the object exits a 2-meter radius around the user.
\end{enumerate}

The NOA device should produce auditory signals for objects based on this ideal warning scheme. For a given sequence, we therefore compute the ideal warning using the full trajectories of all objects in the scene. If at any point in time an object enters the 1-meter safety bubble, the ideal warning starts 3 seconds before that event. There is one caveat: for the synthetic data, ground truth trajectories of the objects are available. However, for the real-world data, we have to rely on the hand segmentation in the images, which is prone to errors. Despite this, we believe this error is small enough to allow for a meaningful comparison of the warning quality across different methods.

\subsection{Metrics}
In this work, we use intersection over union (IoU) as the main metric to evaluate the performance of different collision estimation methods and the warnings they generate.
\begin{equation}
  \text{IoU} =
    \frac{\sum_{f,o}\text{TP}(f,o)}
         {\sum_{f,o}\text{TP}(f,o) + \sum_{f,o}\text{FP}(f,o) + \sum_{f,o}\text{FN}(f,o)}
\label{eq:iou}
\end{equation}
\begin{equation}
  \text{TP}(f,o) =
  \begin{cases}
    1, & W_i(f,o) == 1 \wedge W_m(f,o) == 1\\
    0, & \text{otherwise}
  \end{cases}
\end{equation}
\begin{equation}
  \text{FP}(f,o) =
  \begin{cases}
    1, & W_i(f,o) == 0 \wedge W_m(f,o) == 1\\
    0, & \text{otherwise}
  \end{cases}
\end{equation}
\begin{equation}
  \text{FN}(f,o) =
  \begin{cases}
    1, & W_i(f,o) == 1 \wedge W_m(f,o) == 0\\
    0, & \text{otherwise}
  \end{cases}
\end{equation}

Here, TP, FP and FN are true positives, false positives and false negatives, respectively. These three measures are summed across all frames $f$ and all objects $o$ in all scenes. The basis for computing the single measures is the ideal warning (see Section \ref{sub:ideal_warning}), which is treated as ground truth. A true positive is counted when the ideal warning $W_i$ requires a warning at frame $f$ for object $o$ and the object evaluation method $m$ also leads to a warning $W_m$ for that same object $o$ at that same frame $f$. A false negative occurs when the object evaluation does not lead to a warning, even though the ideal warning requires a warning for object $o$ at frame $f$. Conversely, a false positive is recorded when the object evaluation leads to a warning although the ideal warning does not require one. Again, this is measured per object and per frame.

\subsection{Collision Estimation Baseline Methods}

\begin{figure}[t]
    \centering
    \includegraphics[width=0.96\linewidth]{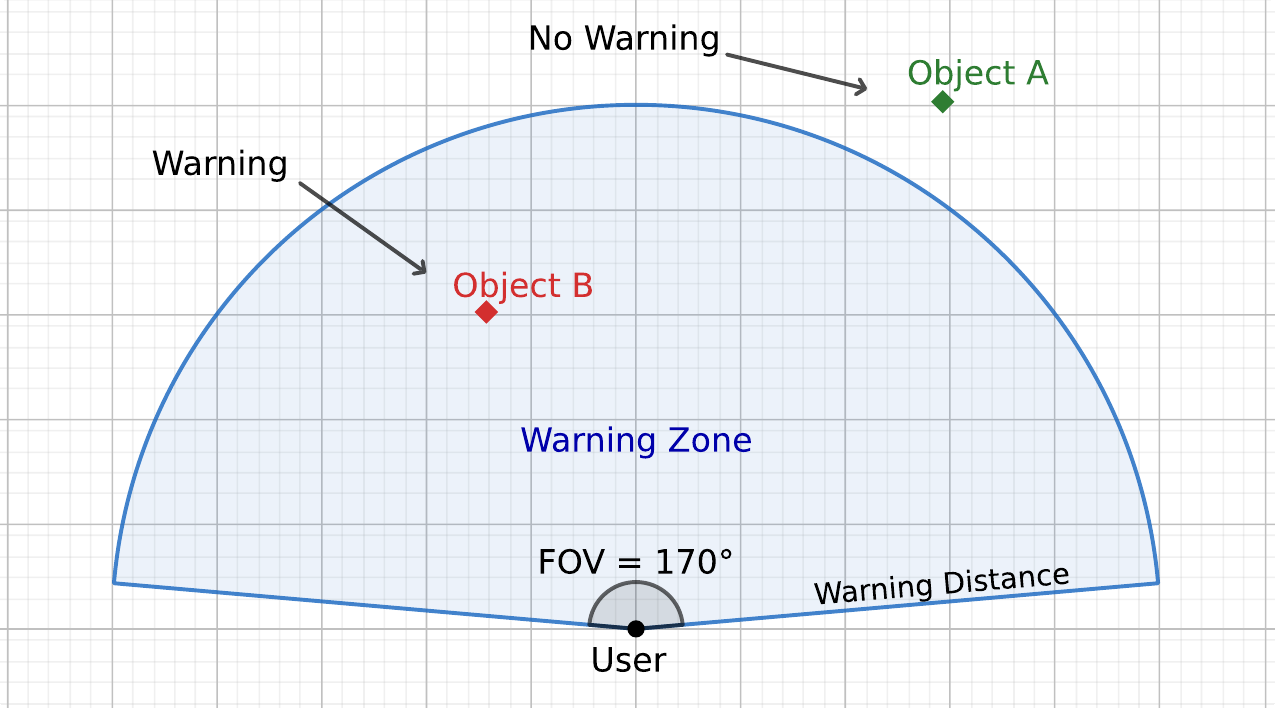}
    \caption{Distance-based baseline collision evaluation: Warnings are triggered for objects that enter the blue safety bubble in front of the user, irrespective of the object's movement direction.}
    \label{fig:baseline}
\end{figure}

The main goal of this work is to evaluate the benefits of integrating state-of-the-art collision estimation methods into blind assist devices. Currently, the NOA device generates warnings based solely on object distance. The collision evaluation is looking at the distance of an object to the user. If this distance falls below a threshold, the object is considered a potential danger and a warning is triggered. Since the cameras only monitor the area in front of the user, warnings are limited to objects within this field of view (see Fig.~\ref{fig:baseline}). A similar distance-based warning was for example applied in \cite{Heimberger_2017} for a parking system.

The distance-based baseline collision evaluation does not consider object movement. As a second baseline, we therefore also implement a time-to-contact approach (TTC) which considers the velocity of the object movement. Examples of such TTC-based warning systems can be found in \cite{Shrinivas_2013} or \cite{Puphal_2023_filter}. Our TTC baseline predicts the user's and object's trajectories using simple linear trajectory predictions and issues a warning if the minimum distance between their trajectories and the corresponding time fall below a threshold.

\section{Risk Model}\label{sec:risk_model}

\begin{figure}[t]
    \centering
    \includegraphics[width=1.0\linewidth]{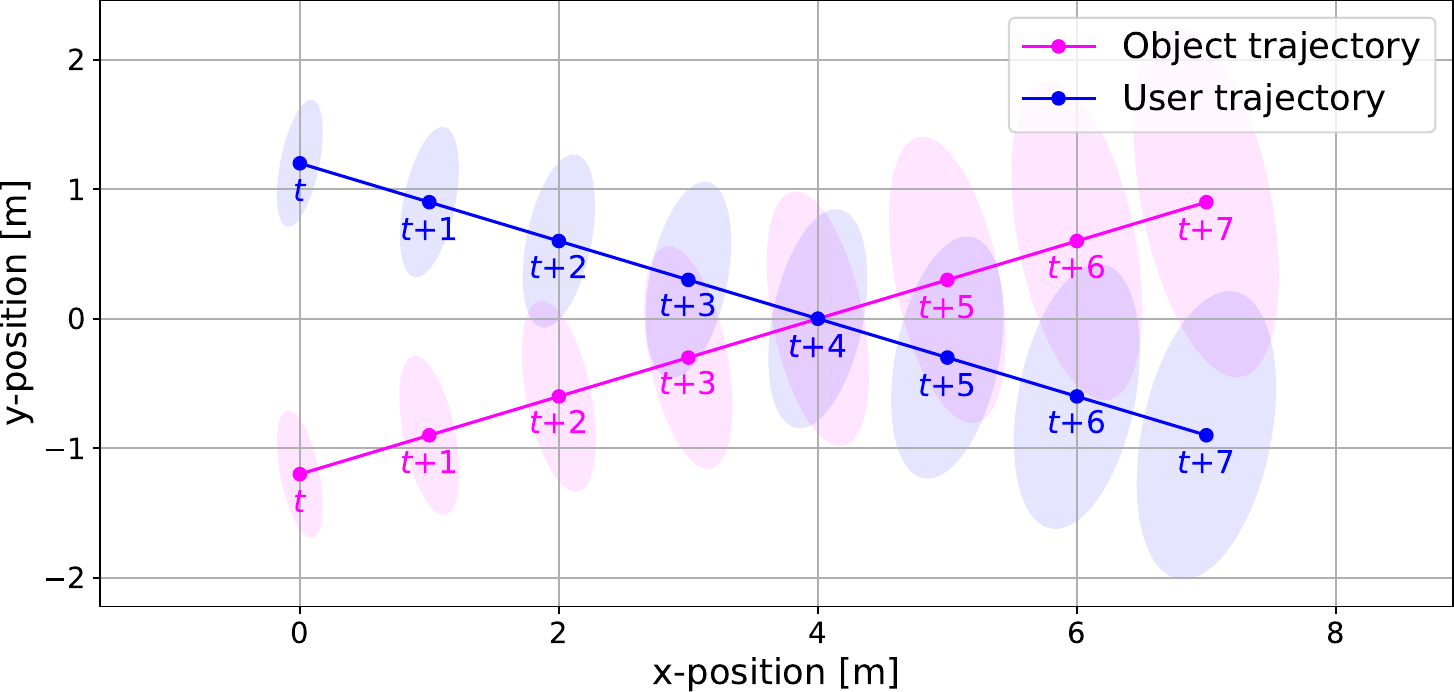}
    \caption{Probabilistic collision collision risk estimation. Future position uncertainties are modeled with Gaussian distributions. In this work, linear object trajectories are predicted from estimated object velocities.}
    \label{fig:linear_trajectory}
\end{figure}

The risk model described in \cite{Puphal_2019} uses Gaussian distributions to model trajectory predictions and uncertainty. In this work, we model linear trajectories for the scene objects with growing Gaussian uncertainty over the predicted time. Fig.~\ref{fig:linear_trajectory} shows an example of modeled Gaussians for both the user and object trajectories.
\par
The collision probability at each predicted timestep can be computed in the risk model as the overlap between two distributions. We calculate the collision probability by taking the integral of the product of two Gaussian functions
\begin{equation}\label{equ:gaus_overlap} 
    P_{\text{coll},i}=\int_{\infty}f_{\text{user}}(x)f_{i}(x)dx,
\end{equation}
where $f_{\text{user}}(x)$ is the Gaussian function for the user over possible positions $x$, and $f_{i}(x)$ is the Gaussian function for object $i$. The Gaussian function $f$ is defined in 1D with a mean position $\mu$ and variance $\sigma^2$ using
\begin{equation}\label{equ:3}
    f(x) = \frac{1}{\sqrt{ 2 \pi \sigma^2 }} \exp\left\{ - \frac{ (x - \mu)^2 } {2 \sigma^2} \right\}.
\end{equation}

\noindent In the risk model, we afterwards account for the dependency between the user and multiple objects by combining the collision risks between the user and all $N$ objects into a total collision probability:
\begin{align}
    P_{\text{coll}} &= P_{\text{coll},i}+ P_{\text{coll},i+1} + ... + P_{\text{coll}, N}(x).
\end{align}
The survival function models the weight assigned to the collision probability at a predicted time $t+s$, starting from the current time $t$ 
\begin{equation}\label{eq:survival_function}
    S(s; t) = \exp\{-\int_{t}^{t+s}(\tau_0^{-1} + \frac{P_{\text{coll}}(s;t)}{\Delta t})ds\},
\end{equation}
where $\tau_0^{-1}$ is the escape rate, representing the likelihood to avoid a collision event, and $\Delta t$ is the duration over which a collision event is considered. The final risk of collision with an object $i$ can finally be computed by taking the integral of the collision probabilities with this object over all predicted timesteps $s$, weighted by the survival function
\begin{equation}
    R_i(t) = \int_{0}^{\infty}S(s;t)\frac{P_{\text{coll},i}(s;t)}{\Delta t}ds.
\end{equation}
For numerical stability, the temporal integration is capped at a fixed prediction horizon $s_{\text{max}}$, which defines the time span of the predicted trajectories and varies depending on the application.

\par
For simplicity, the risk model in Eq. \eqref{equ:3} is described using a 1D Gaussian formulation. In the actual implementation, we employ a 2D Gaussian formulation, accounting for longitudinal and lateral uncertainties that grow over the predicted time (compare Fig. \ref{fig:linear_trajectory}). Unlike the baseline models, this risk model accounts for uncertainties in the object movement, potentially offering greater accuracy in real sensor environments.

\section{Results}\label{sec:results}

\begin{table*}[t]
    \centering
    \caption{Warning accuracy (IoU) for risk model, distance-based warning and time-to-contact (TTC). Three different variants are investigated. \textnormal{Plain}: Warning based on object positions and velocities output from NOA's processing pipeline. \textnormal{Hysteresis}: Applying a hysteresis on generated warning signals to reduce fluctuations. \textnormal{Hysteresis + JPDAF}: Additional application of a post-processing combining a joint probabilistic data association with a particle filter (JPDAF) to NOA's object output.}
    \begin{tabular}{|l|c|c|c|c|c|c|c|c|c|}
        \hline
        & \multicolumn{3}{c|}{Plain} & \multicolumn{3}{c|}{Hysteresis} & \multicolumn{3}{c|}{Hysteresis + JPDAF} \\
        \textbf{Experiment} & \textbf{Risk} & \textbf{TTC} & \textbf{Distance} & \textbf{Risk} & \textbf{TTC} & \textbf{Distance} & \textbf{Risk} & \textbf{TTC} & \textbf{Distance} \\ \hline
        \hline
        \textcolor[HTML]{CCCC00}{1 Scene Object States}      & \textbf{86.78\%} & 83.99\% & 69.08\% & 89.33\% & \textbf{91.39\%} & 76.21\% & 89.33\% & \textbf{91.39\%} & 76.21\% \\ \hline
        \textcolor[HTML]{009900}{2 Annotated Simulated Data} & 73.00\% & \textbf{75.84\%} & 68.90\% & 72.86\% & \textbf{77.24\%} & 72.27\% & \textbf{80.64\%} & 80.21\% & 74.08\% \\ \hline
        \textcolor[HTML]{FF3333}{3 Simulated Data}           & \textbf{66.39\%} & 60.62\% & 57.59\% & \textbf{66.03\%} & 64.86\% & 58.81\% & \textbf{74.36\%} & 66.37\% & 65.90\% \\ \hline
        \textcolor[HTML]{CC6600}{4 Annotated Camera Data}    & 28.46\% & \textbf{48.88\%} & 36.60\% & \textbf{59.57\%} & 54.68\% & 48.91\% & \textbf{70.52\%} & 63.33\% & 61.85\% \\ \hline
        \textcolor[HTML]{0066CC}{5 Camera Data}              & 21.43\% & 17.89\% & \textbf{28.18\%} & 23.58\% & 24.89\% & \textbf{31.35\%} & \textbf{67.00\%} & 51.33\% & 50.90\% \\ \hline
    \end{tabular}
    \vspace{1mm}
    \label{tab:results}
\end{table*}

Using two datasets and five experimental setups (see section~\ref{sub:data}), we compare the risk model with the two baseline approaches based on distance and time-to-contact (TTC). Since the performance of the three approaches also depends on their parameter settings, we optimized the parameters for each algorithm using a genetic algorithm. This way we can compare the best performances of the approaches.

\subsection{Plain}
\label{sub:plain}
First, we investigated the performance of the two baseline approaches and the risk model, for the five experimental setups as shown in Fig.~\ref{fig:pipeline_with_experiments}. The results are shown in Table~\ref{tab:results} in the left block called "Plain". The results give two insights. First, the distance-based approach is inferior to TTC and risk in most of the cases. This is not a surprise because using just the distance completely ignores the dynamics of the objects in the scene. The second observation is that results on the simulated data are much better than the results on real data. While this is not surprising in general, the extend of the difference is. The major reason is the high sensor noise. As the NOA device is tailored for a cost and energy efficient operation, low-cost and energy-friendly sensors are used which, as a downside, exhibit high sensor noise.

The dominating source of noise in the NOA pipeline is position noise. This stems from the noise of the stereo camera's depth measurements. Due to this noise the position of objects is fluctuating over time which in turn leads to fluctuating velocity estimates. Imprecise positions and velocities lead to wrong collision predictions for all three methods. Since the depth map is transformed into a point cloud which is then used to cluster objects, the depth noise also leads to imprecise object clustering. Due to this the objects might be split up or merged from one time step to another and the object IDs might get swapped which will lead to a wrong assignment of collisions to objects. The situation in experimental setup 4 is a bit alleviated by the object masks, provided by the hand segmentation, which stabilize object clustering. However, the performance is well below the simulated data.

\begin{figure*}[t]
    \centering
    \includegraphics[width=0.31\linewidth]{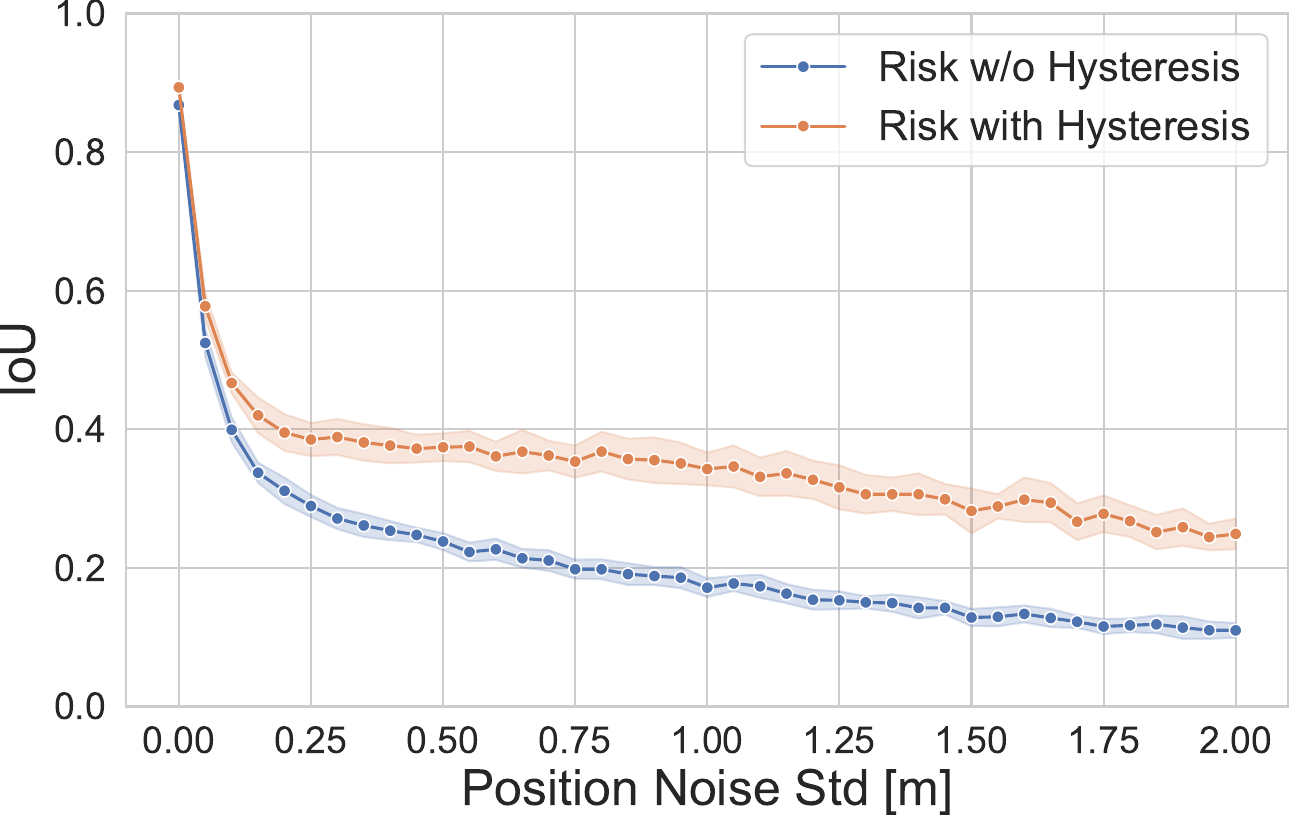}
    \hfill
    \includegraphics[width=0.31\linewidth]{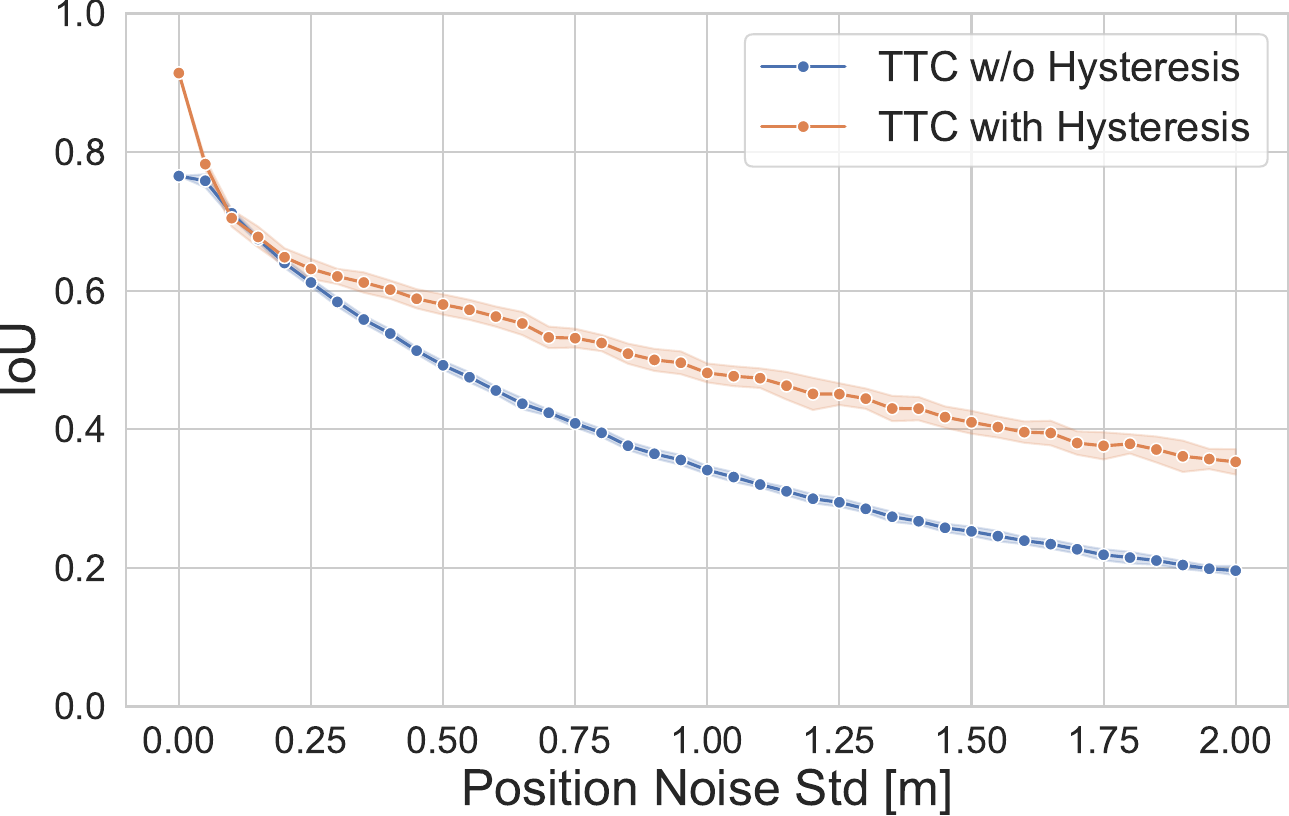}
    \hfill
    \includegraphics[width=0.31\linewidth]{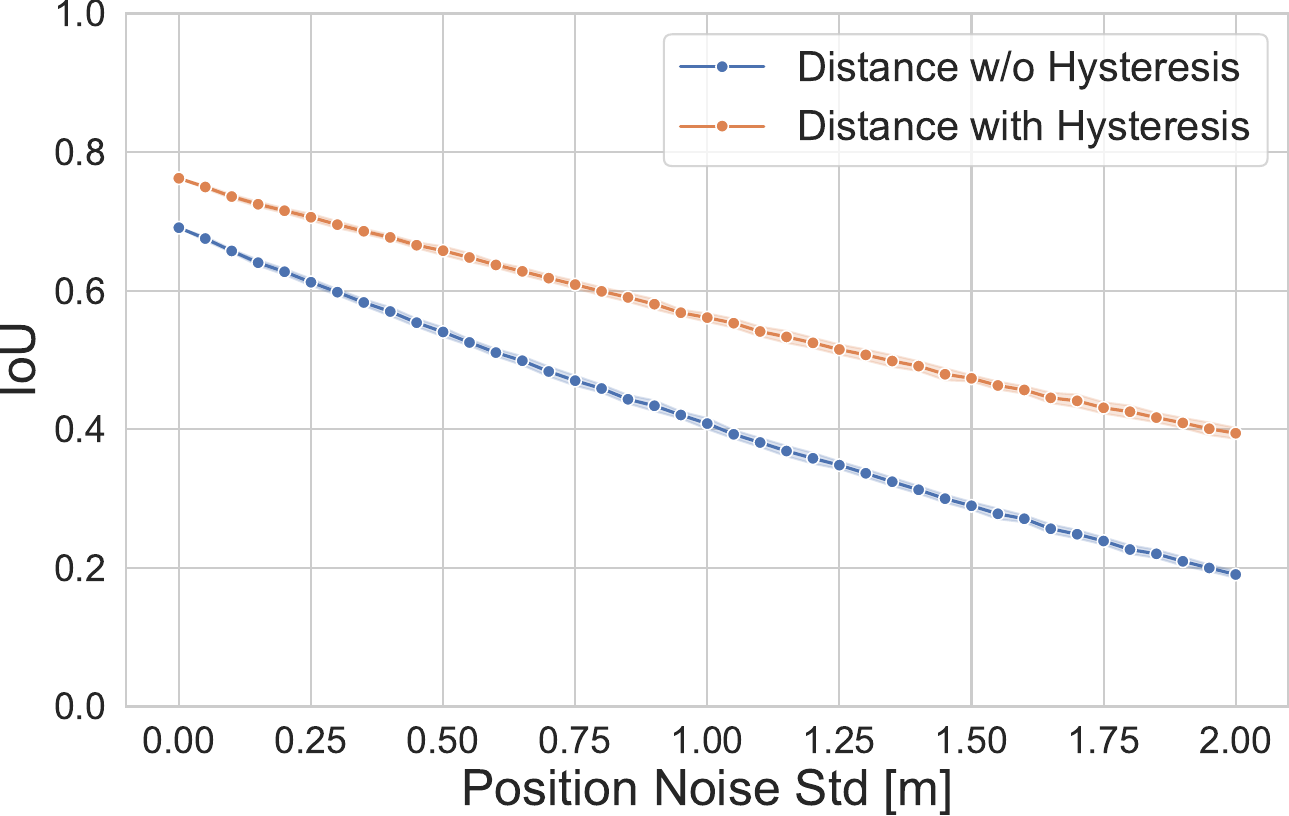}
    \\
    \vspace{5mm}
    \includegraphics[width=0.31\linewidth]{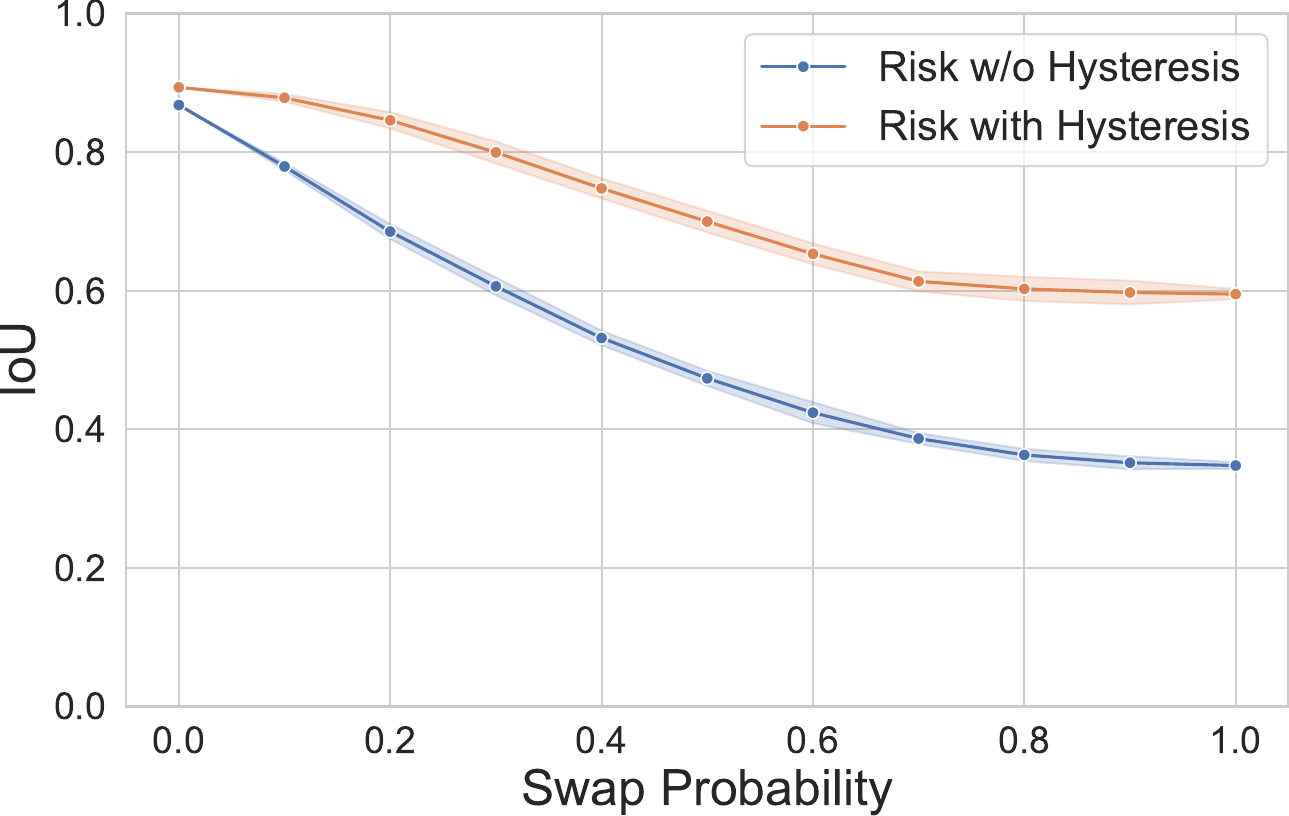}
    \hfill
    \includegraphics[width=0.31\linewidth]{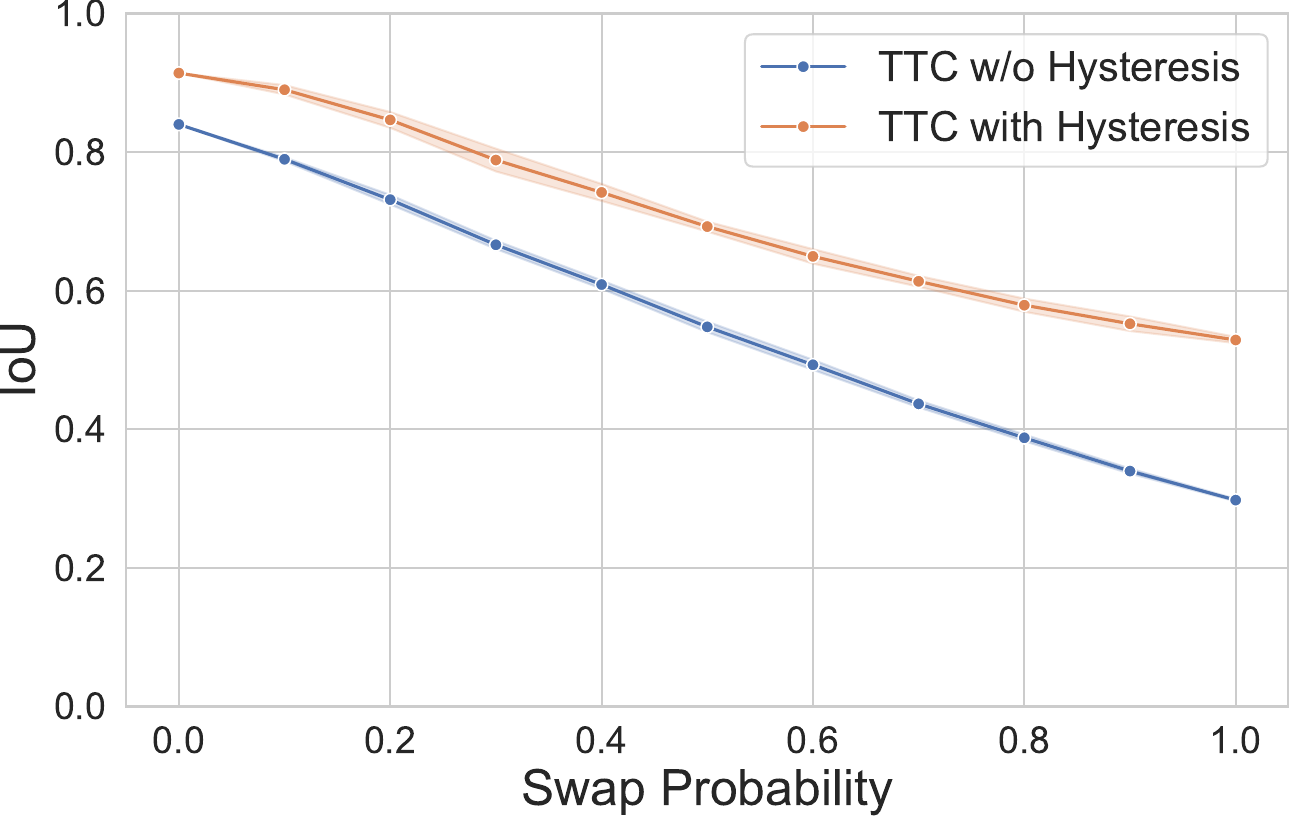}
    \hfill
    \includegraphics[width=0.31\linewidth]{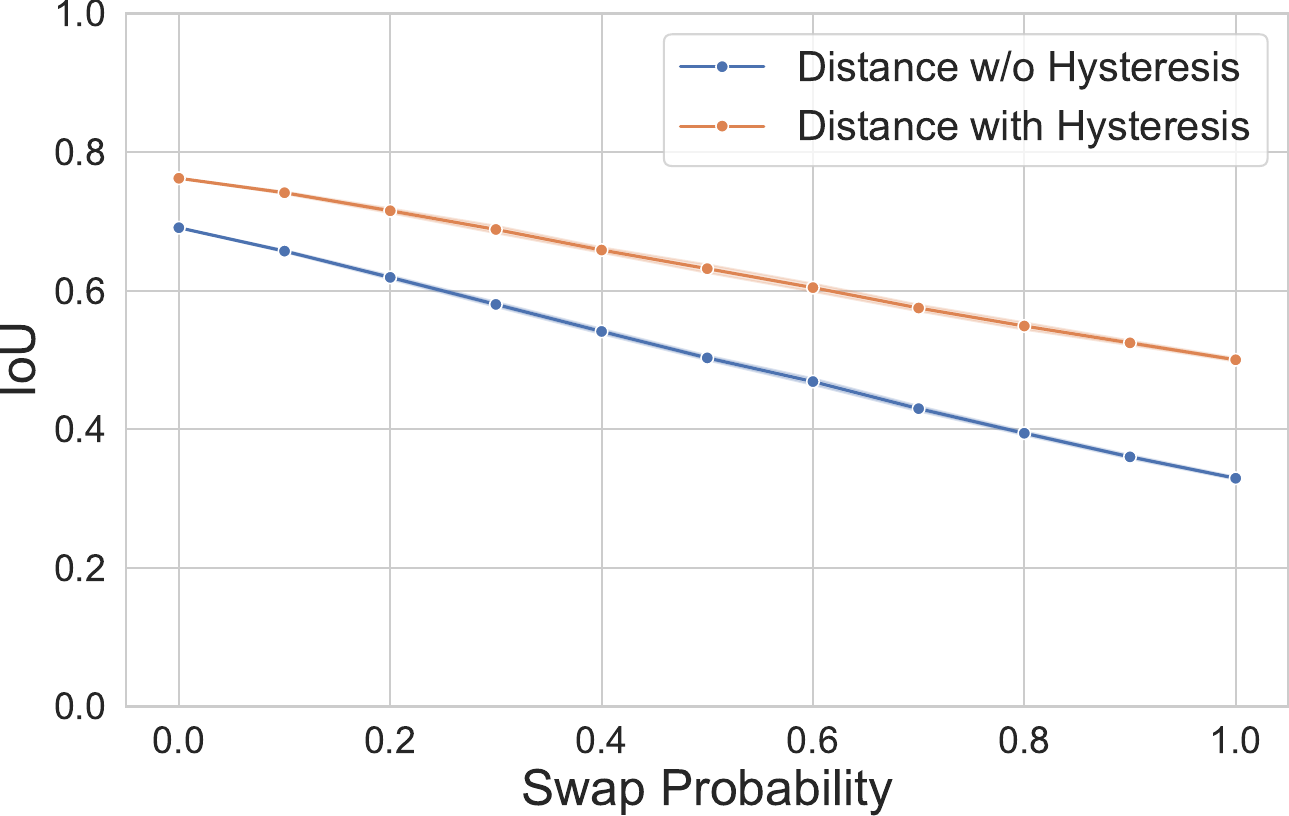}
    \caption{Performance of the risk model (left), TTC (center) and distance-based warning (right) for different levels of object position and ID swap noise. The blue curves show the "Plain" performance of the algorithms, while the red curves show the performance using an additional warning hysteresis.}
    \label{fig:noise_exp}
\end{figure*}

Fig.~\ref{fig:noise_exp} plots the performance of distance-based warning, TTC and risk model for different levels of noise. Here we used the data from experiment 1, i.e., predefined object trajectories. We analyzed the impact of two noise types. First, positional noise by adding Gaussian noise with varying standard deviation $\sigma$ to the object positions. Second, ID swap noise by randomly swapping IDs with a probability $p$. Each noise test was repeated 30 times.
The results show, that all methods struggle with position noise whereas the risk model struggles most and performance drops quickly. ID swap also reduces performance for all methods but the impact is less strong. For the ID swaps, risk model is the least affected of the three methods.

To tackle the noise problems, we added two countermeasures. First, a hysteresis on the warning and, second, a stabilization of the object positions over time using a particle filter alongside a joint probabilistic data association (JPDAF)~\cite{Rezatofighi_2015}.

\subsection{Hysteresis}
\label{sub:hysteresis}

In order to reduce false warnings, we introduce a warning hysteresis. This hysteresis has two thresholds. First, there is a minimal number of consecutive frames, a collision object needs to be detected before an actual warning is triggered. This removes warnings due to spurious detections. Second, there is a minimal number of consecutive frames, no collision is detected before the warning for an object is switched off again. This guarantees continuous warning during temporally short detection dropouts caused by wrong predictions or temporal tracking loss of the objects. The red curves in Fig.~\ref{fig:noise_exp} demonstrate the effectiveness of the warning hysteresis for both position as well as ID swap noise. The results for the data from the five different experimental setups are displayed in the second block of Table~\ref{tab:results} called "Hysteresis".

\begin{figure}
    \centering
    \includegraphics[width=1.0\linewidth]{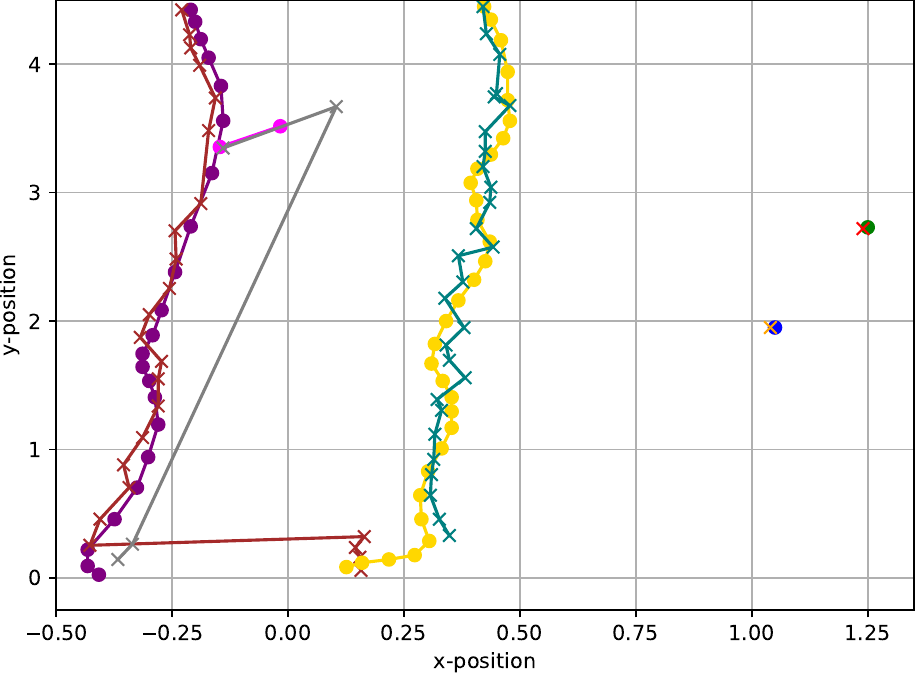}
    \caption{Object tracks generated by NOA (lines with crosses) are prone to swapping. Instead using JDPAF (lines with filled circles) ID swapping is strongly reduced. Each color shows one object track.}
    \label{fig:JPDAF}
\end{figure}

Those results show a clear improvement for all three methods and all experiments. However, while there is quite strong improvement for experiment 4, i.e., real camera data with object annotations, the improvement of real camera data without annotations (experiment 5) is rather low. Real scenes can be very complex and the high noise in depth leads to clustering errors which lead to position errors and swapping of objects during tracking.
Fig.~\ref{fig:JPDAF} shows a typical trajectory example. The brown object is split into two objects (brown and grey) after some frames. Close to the user the cyan object is having a clustering issue and, thus, a jump in position which confused the NOA processing and leads to a swapping of the ID with the brown object.

\subsection{Joint Probabilistic Data Association Filter (JPDAF)}
\label{sub:JPDAF}

To stabilize object tracking and the object IDs, we employ a particle filter (PF) alongside a joint probabilistic data association (JPDA) \cite{Rezatofighi_2015} to the object positions output by the NOA processing pipeline. As the example in Fig.~\ref{fig:JPDAF} shows, the JDPAF leads to much smoother trajectories and more stable IDs. Compare the brown object track from NOA with the purple track after JPDAF and cyan object track from NOA with yellow track after JPDAF.

The numeric results for the risk model and the two baselines with JPDAF are shown in the third block called "Hysteresis + JPDAF" in Table~\ref{tab:results}. They demonstrate the effectiveness of the JPDAF as there is a strong improvement of all three methods for both experiments 4 and 5. For experiment 5, real camera data, the improvement is around 20 percentage points for distance, 26 percentage points for TTC and 42 percentage points for risk so that the distance-based warning reaches 50.90\% accuracy, TTC reaches 51.33\% accuracy and risk reaches 67.00\% accuracy.

After applying the stabilization with JPDAF, warning based on risk clearly outperforms warning based on distance and TTC for real data. However, it also shows that all methods require stable position and velocity data. The experiments here revealed that this is much more difficult to achieve for a wearable blind-assist device as compared to car or robotic systems. Both robots and cars are equipped with better sensors and more processing power. Furthermore, in vehicle scenarios map data can be used to limit the trajectories of recognized vehicles to the street lanes, which strongly improves the position accuracy and also allows for easy prediction of non-linear trajectories \cite{Puphal_2022}. These non-linear trajectories can be exploited by the risk model to improve risk estimation. Hence, it is a target for the future to investigate whether methods for generating non-linear human trajectory predictions can further improve the warning for blind-assistance.

\subsection{Risk Model Parameter Impact}
\label{sub:risk_parameter}

\begin{figure}
    \centering
    \includegraphics[width=0.95\linewidth]{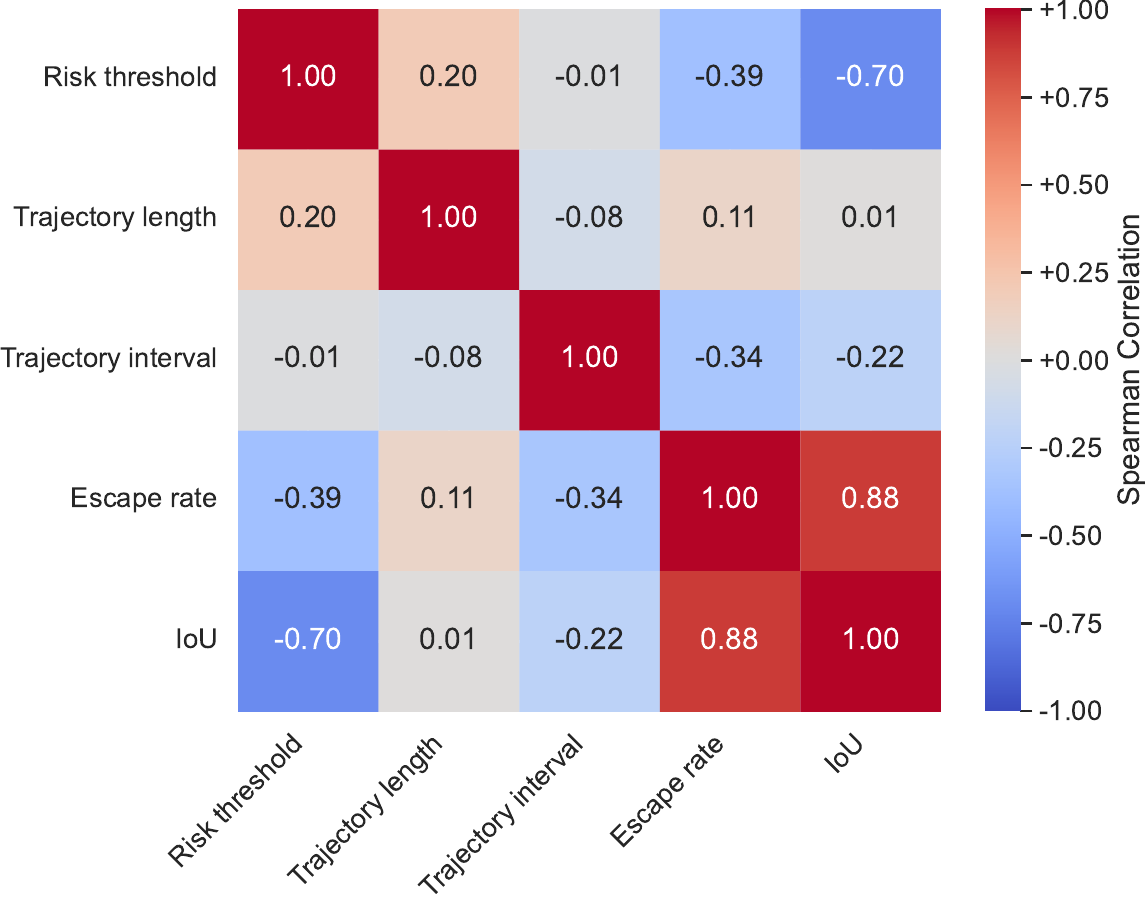}
    \caption{Spearman correlation coefficients among risk model parameters and with the warning accuracy (IoU).}
    \label{fig:corr_standard_risk}
\end{figure}

Another interesting aspect is the impact of the parameters to the warning accuracy in the blind-assistance scenario discussed in this work. To this end, we did a correlation analysis of the risk model parameters, which are {\it risk threshold}, {\it trajectory length},  {\it trajectory interval} and {\it escape rate}. We applied a wide range of values for each parameter and measured how the warning performance changes. Fig.~\ref{fig:corr_standard_risk} depicts a correlation matrix which shows the correlation among the parameters and with respect to the warning accuracy (IoU).

The most impacting parameters are escape rate and risk threshold. The escape rate is the likelihood to avoid a collision event. The risk threshold determines at which risk value an object is considered a collision risk. Trajectory length and trajectory interval define how many second into the future a trajectory is predicted and the sampling rate, respectively. Both have only a minor impact to the quality of warning, provided the length is long enough to capture objects three seconds into the future and the sampling rate is high enough to not miss the object collision. These findings are comparable to observations made for the risk model used in driver assistance \cite{Puphal_2022}.

\section{Conclusion and Outlook}
In this work, we integrated the risk model from vehicle research into the blind assist device NOA and analyzed the potential for improving the collision warning. So far the NOA device warning was mainly based on distance. We compared it to the risk model as well as time-to-collision. Experimental results with simulation as well as recorded real data showed that the risk model is outperforming both warning based on distance and time-to-contact. This confirms previous results for driver assistance systems and autonomous vehicle control. 

Our experiments also revealed that, due to its energy and cost efficient nature, the NOA sensors exhibit strong noise which leads to very noisy object positions and tracking difficulties. We could show that using a simple hysteresis for denoising the warning signals, and a combination of a particle filter with a joint probabilistic data association filter for stabilizing object tracks tremendously improves the quality of all three warning methods. With stable object tracks as input, the risk model has a warning accuracy of 67\% which clearly outperforms distance-based and time-to-contact-based warnings having both an accuracy of 51\%.

So far, only simple linear predictions of object trajectories were used. In the future, we would like to investigate if methods for predicting non-linear methods for pedestrians or crowds can improve the warning based on the risk model. Furthermore, we want to apply more state-of-the-art methods from robotics and intelligent vehicles to blind assist devices to continuously improve the performance of these devices and the lives of people with vision impairment.

\bibliographystyle{IEEEtran}
\bibliography{IEEEabrv,biblio}

\begin{thebibliography}{10}
\providecommand{\url}[1]{#1}
\csname url@samestyle\endcsname
\providecommand{\newblock}{\relax}
\providecommand{\bibinfo}[2]{#2}
\providecommand{\BIBentrySTDinterwordspacing}{\spaceskip=0pt\relax}
\providecommand{\BIBentryALTinterwordstretchfactor}{4}
\providecommand{\BIBentryALTinterwordspacing}{\spaceskip=\fontdimen2\font plus
\BIBentryALTinterwordstretchfactor\fontdimen3\font minus
  \fontdimen4\font\relax}
\providecommand{\BIBforeignlanguage}[2]{{%
\expandafter\ifx\csname l@#1\endcsname\relax
\typeout{** WARNING: IEEEtran.bst: No hyphenation pattern has been}%
\typeout{** loaded for the language `#1'. Using the pattern for}%
\typeout{** the default language instead.}%
\else
\language=\csname l@#1\endcsname
\fi
#2}}
\providecommand{\BIBdecl}{\relax}
\BIBdecl

\bibitem{WHO_2019}
\BIBentryALTinterwordspacing
{World Health Organization}, \emph{World Report on Vision}.\hskip 1em plus
  0.5em minus 0.4em\relax World Health Organization, 2019. [Online]. Available:
  \url{https://www.who.int/publications/i/item/world-report-on-vision}
\BIBentrySTDinterwordspacing

\bibitem{Kahaki_2023}
Z.~R. Kahaki, M.~Karimi, M.~Taherian, and R.~Simi, ``Development and validation
  of a white cane use perceived advantages and disadvantages (wcpad)
  questionnaire,'' \emph{BMC Psychology}, vol.~11, no.~1, p. 253, 2023.

\bibitem{Silverman_2022}
\BIBentryALTinterwordspacing
A.~M. Silverman, C.~R. Rhoads, E.~Bolander, and K.~Bleach, ``The role of guide
  dogs in 2022 and beyond,'' \emph{American Foundation for the Blind}, 2022.
  [Online]. Available: \url{https://afb.org/guidedogs-research}
\BIBentrySTDinterwordspacing

\bibitem{Manduchi_2011}
R.~Manduchi and S.~Kurniawan, ``Mobility-related accidents experienced by
  people with visual impairment,'' \emph{AER Journal: Research and Practice in
  Visual Impairment and Blindness}, vol.~4, no.~2, pp. 44--54, 2011.

\bibitem{Real_2019}
S.~Real and A.~Araujo, ``Navigation systems for the blind and visually
  impaired: Past work, challenges, and open problems,'' \emph{Sensors},
  vol.~19, no.~15, p. 3404, 2019.

\bibitem{Sessner_2022}
J.~Sessner, F.~Dellert, and J.~Franke, ``Multimodal feedback to support the
  navigation of visually impaired people,'' in \emph{2022 IEEE/SICE
  International Symposium on System Integration (SII)}.\hskip 1em plus 0.5em
  minus 0.4em\relax IEEE, 2022, pp. 196--201.

\bibitem{Pittet_2025}
C.~E. Pittet and M.~Fabien, ``{Sensory Cues in Assistive Technologies: a
  Comparative Study of the BuzzClip and NOA},'' \emph{medRxiv}, 2025.

\bibitem{Eggert_2014}
J.~Eggert, ``Predictive risk estimation for intelligent adas functions,'' in
  \emph{17th International IEEE Conference on Intelligent Transportation
  Systems (ITSC)}.\hskip 1em plus 0.5em minus 0.4em\relax IEEE, 2014, pp.
  711--718.

\bibitem{Puphal_2019}
T.~Puphal, M.~Probst, and J.~Eggert, ``Probabilistic uncertainty-aware risk
  spot detector for naturalistic driving,'' \emph{IEEE Transactions on
  Intelligent Vehicles}, vol.~4, no.~3, pp. 406--415, 2019.

\bibitem{Puphal_2018}
T.~Puphal, M.~Probst, Y.~Li, Y.~Sakamoto, and J.~Eggert, ``Optimization of
  velocity ramps with survival analysis for intersection merge-ins,'' in
  \emph{2018 IEEE Intelligent Vehicles Symposium (IV)}, 2018, pp. 1704--1710.

\bibitem{Puphal_2022}
T.~Puphal, B.~Flade, M.~Probst, V.~Willert, J.~Adamy, and J.~Eggert, ``Online
  and predictive warning system for forced lane changes using risk maps,''
  \emph{IEEE Transactions on Intelligent Vehicles}, vol.~7, no.~3, pp.
  616--626, 2022.

\bibitem{Honda_2024}
\BIBentryALTinterwordspacing
{Honda Global}, ``Honda {CI} micro-mobility,'' 2024. [Online]. Available:
  \url{https://global.honda/en/tech/Honda_CI_Micro-mobility/}
\BIBentrySTDinterwordspacing

\bibitem{Michel_2004}
O.~Michel, ``Webots: Professional mobile robot simulation,'' \emph{Journal of
  Advanced Robotics Systems}, vol.~1, no.~1, pp. 39--42, 2004.

\bibitem{Budrionis_2022}
A.~Budrionis, D.~Plikynas, P.~Daniu{\v{s}}is, and A.~Indrulionis,
  ``Smartphone-based computer vision travelling aids for blind and visually
  impaired individuals: A systematic review,'' \emph{Assistive Technology},
  vol.~34, no.~2, pp. 178--194, 2022.

\bibitem{Bai_2017}
J.~Bai, S.~Lian, Z.~Liu, K.~Wang, and D.~Liu, ``Smart guiding glasses for
  visually impaired people in indoor environment,'' \emph{IEEE Transactions on
  Consumer Electronics}, vol.~63, no.~3, pp. 258--266, 2017.

\bibitem{Mukhiddinov_2021}
M.~Mukhiddinov and J.~Cho, ``Smart glass system using deep learning for the
  blind and visually impaired,'' \emph{Electronics}, vol.~10, no.~22, p. 2756,
  2021.

\bibitem{Amore_2023}
F.~Amore, V.~Silvestri, M.~Guidobaldi, M.~Sulfaro, P.~Piscopo, S.~Turco,
  F.~De~Rossi, E.~Rellini, S.~Fortini, S.~Rizzo \emph{et~al.}, ``Efficacy and
  patients’ satisfaction with the orcam myeye device among visually impaired
  people: a multicenter study,'' \emph{Journal of Medical Systems}, vol.~47,
  no.~1, p.~11, 2023.

\bibitem{Khan_2018}
I.~Khan, S.~Khusro, and I.~Ullah, ``Technology-assisted white cane: evaluation
  and future directions,'' \emph{PeerJ}, vol.~6, p. e6058, 2018.

\bibitem{Messaoudi_2020}
M.~D. Messaoudi, B.-A.~J. Menelas, and H.~Mcheick, ``Autonomous smart white
  cane navigation system for indoor usage,'' \emph{Technologies}, vol.~8,
  no.~3, p.~37, 2020.

\bibitem{Mai_2023}
C.~Mai, D.~Xie, L.~Zeng, Z.~Li, Z.~Li, Z.~Qiao, Y.~Qu, G.~Liu, and L.~Li,
  ``Laser sensing and vision sensing smart blind cane: A review,''
  \emph{Sensors}, vol.~23, no.~2, p. 869, 2023.

\bibitem{Bahadir_2012}
S.~K. Bahadir, V.~Koncar, and F.~Kalaoglu, ``Wearable obstacle detection system
  fully integrated to textile structures for visually impaired people,''
  \emph{Sensors and Actuators A: Physical}, vol. 179, pp. 297--311, 2012.

\bibitem{Kruse_2013}
T.~Kruse, A.~K. Pandey, R.~Alami, and A.~Kirsch, ``Human-aware robot
  navigation: A survey,'' \emph{Robotics and Autonomous Systems}, vol.~61,
  no.~12, pp. 1726--1743, 2013.

\bibitem{Gao_2022}
Y.~Gao and C.-M. Huang, ``Evaluation of socially-aware robot navigation,''
  \emph{Frontiers in Robotics and AI}, vol.~8, p. 721317, 2022.

\bibitem{Helbing_1995l}
D.~Helbing and P.~Moln\'ar, ``Social force model for pedestrian dynamics,''
  \emph{Phys. Rev. E}, vol.~51, no.~5, pp. 4282--4286, 1995.

\bibitem{Alahi_2016}
A.~Alahi, K.~Goel, V.~Ramanathan, A.~Robicquet, L.~Fei-Fei, and S.~Savarese,
  ``Social lstm: Human trajectory prediction in crowded spaces,'' in
  \emph{Proceedings of the IEEE conference on computer vision and pattern
  recognition}, 2016, pp. 961--971.

\bibitem{Aoude_2013}
G.~S. Aoude, B.~D. Luders, J.~M. Joseph, N.~Roy, and J.~P. How,
  ``Probabilistically safe motion planning to avoid dynamic obstacles with
  uncertain motion patterns,'' \emph{Autonomous Robots}, vol.~35, pp. 51--76,
  2013.

\bibitem{Tai_2017}
L.~Tai, G.~Paolo, and M.~Liu, ``Virtual-to-real deep reinforcement learning:
  Continuous control of mobile robots for mapless navigation,'' in \emph{2017
  IEEE/RSJ international conference on intelligent robots and systems
  (IROS)}.\hskip 1em plus 0.5em minus 0.4em\relax IEEE, 2017, pp. 31--36.

\bibitem{Long_2018}
P.~Long, T.~Fan, X.~Liao, W.~Liu, H.~Zhang, and J.~Pan, ``Towards optimally
  decentralized multi-robot collision avoidance via deep reinforcement
  learning,'' in \emph{2018 IEEE international conference on robotics and
  automation (ICRA)}.\hskip 1em plus 0.5em minus 0.4em\relax IEEE, 2018, pp.
  6252--6259.

\bibitem{Everett_2018}
M.~Everett, Y.~F. Chen, and J.~P. How, ``Motion planning among dynamic,
  decision-making agents with deep reinforcement learning,'' in \emph{2018
  IEEE/RSJ International Conference on Intelligent Robots and Systems
  (IROS)}.\hskip 1em plus 0.5em minus 0.4em\relax IEEE, 2018, pp. 3052--3059.

\bibitem{Chen_2019}
C.~Chen, Y.~Liu, S.~Kreiss, and A.~Alahi, ``Crowd-robot interaction:
  Crowd-aware robot navigation with attention-based deep reinforcement
  learning,'' in \emph{2019 international conference on robotics and automation
  (ICRA)}.\hskip 1em plus 0.5em minus 0.4em\relax IEEE, 2019, pp. 6015--6022.

\bibitem{Bliss_2003}
J.~P. Bliss and S.~A. Acton, ``Alarm mistrust in automobiles: how collision
  alarm reliability affects driving,'' \emph{Applied ergonomics}, vol.~34,
  no.~6, pp. 499--509, 2003.

\bibitem{Biondi_2018}
F.~N. Biondi, D.~Getty, M.~M. McCarty, R.~M. Goethe, J.~M. Cooper, and D.~L.
  Strayer, ``The challenge of advanced driver assistance systems assessment: A
  scale for the assessment of the human--machine interface of advanced driver
  assistance technology,'' \emph{Transportation research record}, vol. 2672,
  no.~37, pp. 113--122, 2018.

\bibitem{Hasenjager_2019}
M.~Hasenj{\"a}ger, M.~Heckmann, and H.~Wersing, ``A survey of personalization
  for advanced driver assistance systems,'' \emph{IEEE Transactions on
  Intelligent Vehicles}, vol.~5, no.~2, pp. 335--344, 2019.

\bibitem{Heimberger_2017}
M.~Heimberger, J.~Horgan, C.~Hughes, J.~McDonald, and S.~Yogamani, ``Computer
  vision in automated parking systems: Design, implementation and challenges,''
  \emph{Image and Vision Computing}, vol.~68, pp. 88--101, 2017.

\bibitem{Shrinivas_2013}
S.~Pundlik, M.~Tomasi, and G.~Luo, ``Collision detection for visually impaired
  from a body-mounted camera,'' in \emph{2013 IEEE Conference on Computer
  Vision and Pattern Recognition (CVPR) Workshops}, 2013, pp. 41--47.

\bibitem{Puphal_2023_filter}
T.~Puphal, R.~Wenzel, B.~Flade, M.~Probst, and J.~Eggert, ``Importance
  filtering with risk models for complex driving situations,'' in \emph{2022
  7th International Conference on Robotics and Automation Engineering (ICRAE)},
  2022, pp. 376--383.

\bibitem{Rezatofighi_2015}
S.~H. Rezatofighi, A.~Milan, Z.~Zhang, Q.~Shi, A.~Dick, and I.~Reid, ``Joint
  probabilistic data association revisited,'' in \emph{Proceedings of the IEEE
  international conference on computer vision}, 2015, pp. 3047--3055.

\end{thebibliography}

\end{document}